\documentclass{article}

\usepackage{arxiv}
\usepackage{amsmath}
\usepackage{bbm}
\usepackage[utf8]{inputenc}
\usepackage[T1]{fontenc}   
\usepackage{hyperref}      
\usepackage{url}          
\usepackage{booktabs}      
\usepackage{amsfonts}    
\usepackage{nicefrac}      
\usepackage{microtype}    
\usepackage{lipsum}
\usepackage{graphicx}
\usepackage[table,xcdraw]{xcolor}
\usepackage[numbers]{natbib}
\usepackage{doi}
\usepackage{todonotes}
\usepackage{svg}
\usepackage{relsize}
\usepackage{booktabs}       
\usepackage{wrapfig}

\graphicspath{ {./images/} }
\title{SurvSHAP(t): Time-dependent explanations\\ of machine learning survival models}

\author{
 Mateusz Krzyziński \\
Faculty of~Mathematics and~Information Science \\ 
Warsaw University of~Technology \\
  \texttt{mateusz.krzyzinski.stud@pw.edu.pl} \\
   \And
 Mikołaj Spytek\\
Faculty of~Mathematics and~Information Science \\ 
Warsaw University of~Technology 
  \And
 Hubert Baniecki \\
Faculty of~Mathematics and~Information Science \\ 
Warsaw University of~Technology 
\And
 Przemysław Biecek \\
Faculty of~Mathematics and~Information Science \\ 
Warsaw University of~Technology 
}

\begin{document}
\maketitle
\begin{abstract}

Machine and deep learning survival models demonstrate similar or even improved time-to-event prediction capabilities compared to classical statistical learning methods yet are too complex to~be interpreted by humans. Several model-agnostic explanations are available to overcome this issue; however, none directly explain the survival function prediction. In this paper, we introduce SurvSHAP(t), the~first time-dependent explanation that allows for interpreting survival black-box models. It is based on SHapley Additive exPlanations with solid theoretical foundations and a broad adoption among machine learning practitioners. The proposed methods aim to enhance precision diagnostics and~support domain experts in making decisions. Experiments on synthetic and medical data confirm that SurvSHAP(t) can detect variables with a time-dependent effect, and its aggregation is a better determinant of the importance of variables for a prediction than SurvLIME. SurvSHAP(t) is model-agnostic and can be applied to all models with functional output. We provide an accessible implementation of~time-dependent explanations in Python at \url{https://github.com/MI2DataLab/survshap}.

\end{abstract}


\section{Introduction}
\label{sec:introduction}

Machine learning has been gaining popularity in practical applications to solve various problems. Especially in the medical domain, its capabilities prove highly advantageous. Presumably, every medical professional will work with AI technology in the future, particularly deep learning \cite{topolnature}. Unfortunately, many machine learning models, especially complex ones such as deep neural networks~\cite{deepsurv, deephit, deepomix}, are considered black-box models, i.e., it is not possible to know directly what influences their prediction internally~\cite{ema}. Such knowledge proves helpful for explaining and examining the model of interest.
 
It enables to verify that the predictions are made on the same basis that human domain experts would make them and increases trust in model predictions. The fact that black-box models do not come with readily available explanations has been a hindrance in their widespread adoption, as in many areas, including medicine, it is crucial to know what affects the model output. Therefore, the need for research on explanation methods of such complex models has been postulated~\cite{ml-clinical-dss, xai42}. However, often worse-performing but more common and well-established models are still preferred in many settings~\cite{occams-razor}.

This phenomenon is evident in the field of survival analysis. In this area, the most common are parametric or semi-parametric statistical models with the semi-interpretable Cox Proportional Hazards model~\cite{cox} (CPH or Cox model in short) at the forefront~\cite{cox-popular-2, cox-popular-1}. However, due to the limitations of CPH in modeling complex dependencies and its strong assumptions that are often not met~\cite{ph-assumption}, machine learning models, which are more flexible, have become used in practice, both in healthcare~\cite{ml-uveal-melanoma, survival-ml-application-2, survival-ml-application-3} and in other fields~\cite{kbs-survival-ml-application, survival-ml-application-4}. The growing popularity of this class of models motivates the need to develop methods for explaining their predictions.

To the best of our knowledge, there are few existing post-hoc explanation methods specific to complex survival models. These are counterfactual explanations~\cite{counterfactuals}, a technique based on the use of neural additive models called SurvNAM~\cite{survnam}, and different versions of SurvLIME~\cite{survlime-ks, survlime} -- an approach inspired by LIME~\cite{lime}. However, in none of these methods, the time dimension, crucial for predicting the survival conditional probability distribution, is included in the final explanation. 

\begin{figure}[t]
  \centering
  \includegraphics[width=\textwidth]{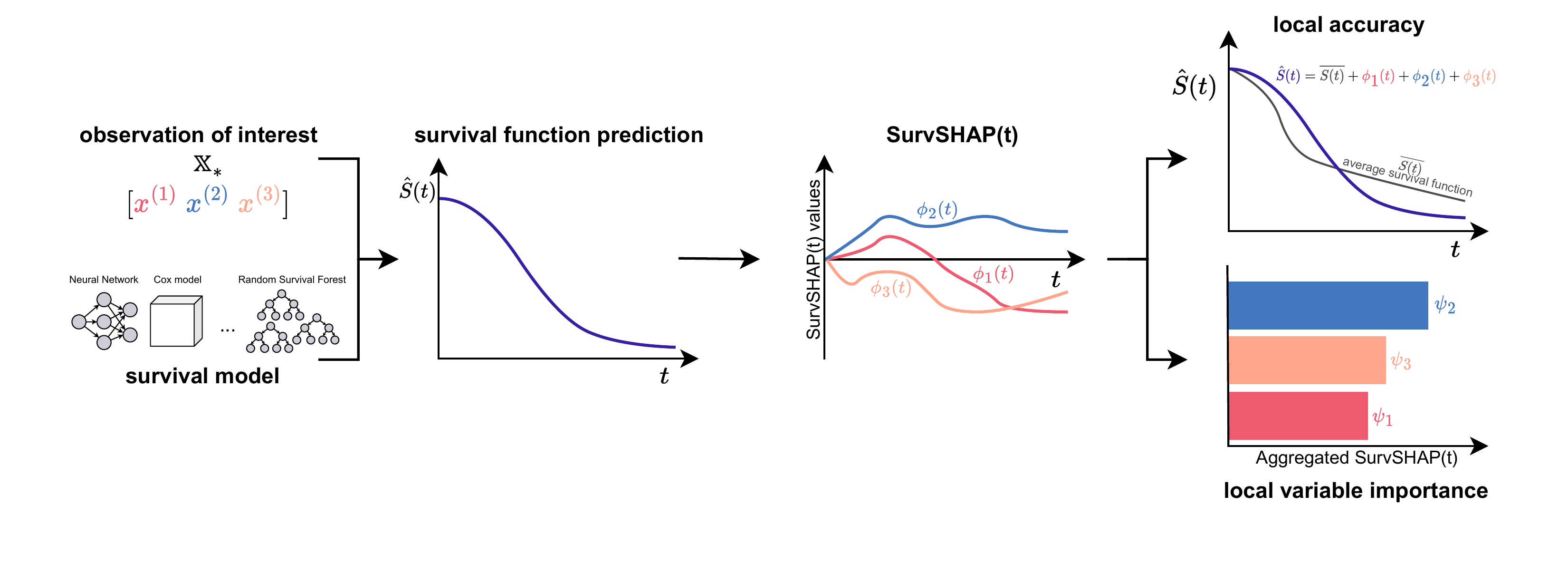}
  \caption{SurvSHAP(t) allows for time-dependent explainability of any survival model predictions. SurvSHAP(t) values add up to the survival function predicted by the model and aggregated over time can be treated as a local importance ranking of variables.}
  \label{fig:diagram}
\end{figure}

Thus, we aim to extend the available techniques by presenting the new method called SurvSHAP(t) that fills this gap (see diagram in Figure~\ref{fig:diagram}). The proposed solution generalizes SHapley Additive exPlanations (SHAP)~\cite{shap} to survival models. However, the method can be applied to any model with functional output. In the survival analysis setting, the time dimension is used in the proposed explanations -- they provide an insight into how each variable influences the model’s response (survival function) at each time point. We refer to this property as \emph{time-dependent} explainability. 
The contributions of this paper can be summarized as follows:
\begin{enumerate}
    \item We introduce SurvSHAP(t) -- the first time-dependent explanation that allows for interpreting any survival model with functional output. It is based on SHAP with solid theoretical foundations and a broad adoption among machine learning practitioners.
    \item We prove that SurvSHAP(t) meets the local accuracy property and accurately explains the model predictions in the form of a survival function, describing variable contributions across the entire analyzed time range. The conducted experiments confirm that SurvSHAP(t) is able to detect variables with a time-dependent effect, and its aggregation is a better determinant of the importance of variables for a prediction than SurvLIME.
    \item We provide an accessible implementation of both SurvSHAP(t) and SurvLIME in Python. Source code for both methods and the data generation is available on GitHub at \url{https://github.com/MI2DataLab/survshap}.
\end{enumerate}

\newpage
\section{Related work}
\label{sec:related-work}

\paragraph{Machine learning survival models.}

The most fundamental and frequently used approach to survival tasks is applying the Cox Proportional Hazards model, which is a semi-parametric model. It has limitations such as a degradation of performance when working with high dimensional data or correlated variables. A strong proportional hazards assumption has highlighted the need for developing methods based on classical machine learning techniques appropriately adapted to the censored data~\cite{ml-survival-survey}.

One popular survival machine learning model, which is used in our experiments, is Random Survival Forest (RSF)~\cite{rsf}. The most common splitting rule for forming individual decision trees is the log-rank splitting rule based on the log-rank test~\cite{survivaltree}. The single survival tree prediction for an individual is a function computed for all individuals in the same tree terminal node; most often a cumulative hazard function estimated using the Nelson-Aalen estimator~\cite{aalen, nelson}. The entire prediction of RSF is the function averaged over all trees, which makes them able to predict complicated survival functions. Another class of models applied to survival analysis is Gradient Boosting Machines (GBM)~\cite{boosting}. A GBM performs a greedy stage-wise process with the objective of optimizing a selected function. In the basic case, it maximizes the log-partial likelihood known from the Cox model. The likelihood can also be substituted with a differentiable approximation of the concordance index -- a metric used to evaluate the survival models' performance~\cite{sgb-concordance}. Implementations of these models are accessible through open-source software like \texttt{scikit-survival}~\cite{sksurv} Python package. In R, many popular packages can be adapted for survival predictions, including \texttt{randomForestSRC}~\cite{randomForestSRC} and \texttt{gbm}~\cite{gbm}. 

Early adaptations of neural networks to survival analysis were proposed in the 1990s, e.g., a non-linear proportional hazards model trained using the partial likelihood function~\cite{faraggi-simon}. A similar idea is also used in more recent approaches to survival prediction, such as Cox-nnet~\cite{ching2018coxnnet} -- the main difference is the modification of the loss function. Modern deep learning models, such as DeepSurv~\cite{deepsurv} and Cox-Time~\cite{pycox}, also often model a prognostic index corresponding to the linear predictor from the Cox model. However, there are also architectures that return more flexible predictions without relying on the Cox model~\cite{rnn-surv, deephit, dnnsurv}. Another notable deep learning approach is DeepOmix, which has been proposed as an interpretable model for multi-omics data~\cite{deepomix}. 

Unfortunately, relatively few survival deep learning models have open-source implementations. Some of the solutions are available in \texttt{pycox}~\cite{pycox} and \texttt{auton-survival}~\cite{auton-survival} Python packages, or \texttt{survivalmodels}~\cite{survivalmodels} R package.

\paragraph{Explanations of machine learning models.}

\emph{Interpretability} has many definitions, and one widely-used formulation is \emph{the degree to which a human can consistently understand model predictions}~\cite{kim-interpretability}. Many complex machine and deep learning models are not directly interpretable, which leads to the development of eXplainable AI (XAI)~\cite{holzinger2022xaioverview}. Explanation of machine learning models can be divided into two categories: local methods, which are used to explain the model’s predictions for a particular observation, and global ones, which provide information about the overall behavior of the model. The most widespread model-agnostic local explanations are additive variable attributions: Local Interpretable Model-agnostic Explanations (LIME)~\cite{lime} and SHapley Adaptive exPlanations (SHAP)~\cite{shap}. 

LIME~\cite{lime} uses a local interpretable surrogate model fitted to the vicinity (a new dataset generated artificially) of an individual observation. For tasks of regression and classification, linear and logistic regression models are used, whose coefficients have clear interpretations. 

SHAP~\cite{shap} is based on the Shapley value framework from game theory~\cite{Shapley}, which was introduced into interpreting machine learning predictions in~\cite{strumbelj1, strumbelj2}. Shapley values characterize how much each variable influences the model prediction in relation to the baseline average. Attributions are calculated as a mean change to the prediction after adding the examined variable to each possible subset of the model's variables. KernelSHAP~\cite{shap} is proposed as an exact estimation of Shapley values inspired by LIME. Exact KernelSHAP has two main limitations: high computation cost and producing misleading explanations when variables are dependent. The first can be improved when explaining tree-ensemble models like random forests with the efficient TreeSHAP algorithm~\cite{treeshap}. The second can be overcome by a more thoughtful generation of neighbourhood samples under the curse of dimensionality~\cite{shapr} and the use of variational encoders to model feature dependences~\cite{vaeac}. Many more explanations based on the SHAP framework like TimeSHAP~\cite{timeshap} and FastSHAP~\cite{fastshap} advance our understanding of complex learning algorithms.

Alike implementations of survival machine learning models, explanations for classification and regression models are available through open-source software both in Python (\texttt{dalex}~\cite{dalex-python}, \texttt{shap}~\cite{shap}) and R (\texttt{DALEX}~\cite{dalex-r}, \texttt{shapr}~\cite{shapr-implementation}).

\paragraph{Explanations of machine learning survival models.}

Some explanations of predictive models for the more standard regression and classification tasks can be adapted for survival models with the use of single-point risk predictions or aggregations of survival functions \cite{moncada2021survshap}. However, this leads to the loss of information contained in the survival distribution, especially in the case of complex models algorithms modeling flexible survival functions. In contrast, SurvSHAP(t) provides explanations of the whole distribution in the form of the survival function. 

To overcome the mentioned shortcomings, \cite{survlime} proposes to adapt LIME into the SurvLIME method by using another object describing survival distribution -- the cumulative hazard function (CHF) -- as the basis for calculations. The Cox Proportional Hazards model is used as a surrogate model, whose coefficients are fitted by optimizing a loss function based on the distances between CHFs predicted by the local surrogate model and the black-box model. Like in LIME, the optimization problem is based on a sample of weighted observations from the local area around the point of interest. SurvLIME uses the $L^2$ metric to calculate the distance between functions. SurvLIME-KS~\cite{survlime-ks} is an extension to the method that uses Kolmogorov-Smirnov bounds for constructing sets of predicted CHFs, which helps to robustify the explanation. However, these methods take into account the distribution only in the computation phase, and the obtained results are the coefficients of the Cox Proportional Hazards model (single values). SurvSHAP(t) extends the dependency on the distribution by providing time-dependent explainability, i.e., it returns an explanation for the entire support of the considered distribution (time range) while being able to aggregate it into meaningful single values.

Moreover, counterfactual explanations for survival analysis models have been proposed, in which the survival function is used to find the counterfactual~\cite{counterfactuals}. Specifically, the difference between the two survival functions, for the original point of interest and the counterfactual, is based on the mean time-to-event distance for the set of observations. This method falls under a different explanation methods category than SurvSHAP(t) as it does not return variable attributions.

Despite the existence of several explanation methods, further development is needed in the field of XAI for survival analysis, and certain problems are apparent. Firstly, none of the known approaches supply time-dependent explanations. Another major deficiency is the lack  of publicly available implementations of the methods described in the literature. Therefore, we present SurvSHAP(t) with the expected properties of the survival model explanation along with its implementation.

\section{Preliminaries}
\label{sec:preliminaries}
We first define the notation needed to later introduce time-dependent explanations in Section~\ref{sec:methods}. 
\subsection{Mathematical background of survival analysis}
Survival analysis deals with tasks based on censored data. That means we have incomplete information about an individual’s survival time for a part of the population from the dataset. Usually, the case of right censoring is considered -- during the study, the event of interest, e.g., a patient's death, is not observed for some part of the population. That means the observed event time is less than or equal to the actual survival time.

Mathematically, a given instance $i$ is represented as a triplet  $(\textbf{x}_i, y_i, \delta_i)$, where $\textbf{x}_i = [x_i^{1}, x_i^{2}, \ldots, x_i^{p}] \in \mathbb{R}^p$ indicates the variables vector; $\delta_i$ is the indicator of event of interest's occurrence; and $y_i$ stands for the observed time (either survival time $T_i$ when $\delta_i = 1$ or censoring time $C_i$ when $\delta_i = 0$). Thus, one should acknowledge that $T_i$ is a latent value for censored observations. The primary objective of survival analysis is the estimation of $T_j$ for an instance $j$ with variables vector $\textbf{x}_j$. Most often, instead of predicting a single time moment, a certain function of time is the output. 

The first key object is the survival function~\eqref{survival-function} which describes the probability of an individual surviving until time $t$ without experiencing the event, i.e.,

\begin{align}
S(t) = \mathbb{P}(T>t) = 1 - \mathbb{P}(T\leq t). \label{survival-function}
\end{align}

Another fundamental concept is the hazard function~\eqref{hazard-function} that can be interpreted as the conditional failure rate in a short (infinitesimal) time interval, provided that the event has not occurred by time $t$, which is defined as

\begin{align}
    h(t) =  \lim_{t\xrightarrow{}0} \frac{\mathbb{P}(t \leq T < t + \Delta t \, | \, T \geq t)}{\Delta t} = \frac{f(t)}{S(t)},  \label{hazard-function}
\end{align}

where $f(t) = - \frac{\mathrm{d}S(t)}{\mathrm{d}t}$ is the event of interest's density function. 

Therefore, the survival function is connected with the hazard function and can be rewritten as 
\begin{align}
    S(t) = \exp(-H(t)), \label{relationship-S-H}
\end{align}
where $H(t) = \int_0^t h(s)\,\mathrm{d}s$ is called the cumulative hazard function.  

Moreover, based on the survival function, selecting the appropriate aggregation makes it possible to obtain other predictions of the type of the single values, e.g., time until an event of interest, risk score \cite{sonabend-phd}. 

\subsection{Variable importance ranking in CPH and SurvLIME}
\label{sec:CPH-ranking}

The coefficients of a Cox Proportional Hazards model can be used to rank the relative importance of variables for a given prediction. We achieve this by comparing the absolute values of the coefficients multiplied by the values of its variables $|x^{(d)}\cdot b^{(d)}|$ for each variable $d$ -- higher values indicate higher importance. Note that simply comparing the values of $\textbf{b}$ coefficients does not inform the end-user about the importance of variables as they can be of vastly different scales. The interpretation of a coefficient $b^{(d)}$ on its own is that an increase of the variable $x^{(d)}$ by one unit indicates that the hazard rate for the given observation will be $\exp(b^{(d)})$ times higher than the actual value.

We later use this method of ranking variables when comparing SurvLIME to SurvSHAP(t) in Section \ref{sec:compare-survlime}, as the SurvLIME explanation takes the form of Cox model coefficients.

\section{Time-dependent explanations of machine learning survival models}
\label{sec:methods}

Let $\mathbb{D} = \{(\textbf{x}_i, y_i, \delta_i): i=1, 2, \ldots, n\}$ be the survival dataset used for training the black-box model. Moreover, assume that $t_1 < t_2 < \ldots < t_m$ are distinct times to event of interest from the set $\{y_i: \delta_i = 1; i=1, 2, \ldots, n \}$. For each individual described by a variables vector $\textbf{x}$, the model returns the individual's survival distribution $\hat{S}(t,\textbf{x})$ (i.e., the distribution of the event of interest occurring over $\mathbb{R}_{\geq0}$). The returned object is the survival function as it uniquely determines the distribution. Note that most often, the value of $\hat{S}(t,\textbf{x})$ is known for all $t \in \{t_1, \ldots, t_m\}$ and for the remaining points, interpolations or step functions are used.

The idea behind the proposed SurvSHAP(t) method is to use the survival function not only to compute an explanation but also to present its results. For this purpose, for the observation of interest $\textbf{x}_*$ at any selected time point $t$ the algorithm assigns an attribution (importance value) $\phi_{t}(\textbf{x}_*, d)$ to the value of each variable $x^{(d)}$, $d \in \{1, 2, \ldots, p\}$, included in the model. In this way, the SurvSHAP(t) functions $[\phi_{t_1}(\textbf{x}_*, d), \phi_{t_2}(\textbf{x}_*, d), \ldots, \phi_{t_{m}}(\textbf{x}_*, d)]$ are generated for every predictor $d$. These functions describe the time-dependent influence of variables on the prediction of the model.

We implement SurvSHAP(t) estimation in two ways, which are based on regression and classification approaches. The first is the Shapley sampling values algorithm. Let $e_{t, \textbf{x}_*}^D = \mathbb{E}[\hat{S}(t,\textbf{x})|\textbf{x}^D = \textbf{x}_*^D  ]$ be the expected value for a conditional distribution where conditioning applies to all variables from the set $D$.

The contribution of predictor $d$ in time point $t$ is calculated as
\begin{align}
\phi_{t}(\textbf{x}_*, d) = \frac{1}{|\Pi|} \sum_{\pi \in \Pi} e_{t, \textbf{x}_*}^{\mathrm{before}(\pi, d) \cup \{d\}} - e_{t, \textbf{x}_*}^{\mathrm{before}(\pi, d)}, 
\end{align}
where $\Pi$ is a set of all permutations of $p$ variables and $\mathrm{before}(\pi, d)$ denotes a subset of predictors that are before $d$ in the ordering $\pi \in \Pi$. For easier comparison between different models and time points, this value can be normalized to obtain values on a common scale (from -1 to 1) according to the formula 

\begin{align}
    \phi^*_{t}(\textbf{x}_*, d) = \frac{\phi_{t}(\textbf{x}_*, d)}{\sum_{j=1}^{p}|\phi_{t}(\textbf{x}_*, j)|}. \label{SHAP-normalization}
\end{align}

It should be noted that thanks to the property that the expected value of the random vector is the vector of the expected values, this operation can be vectorized and performed simultaneously for all selected time points. However, due to the computational cost, permutation sampling is used in the case of high-dimensional models.

Another way to calculate SurvSHAP(t) faster is to use the Shapley kernel~\cite{shap} and weighted linear regression with functional responses and scalar variables~\cite{faraway-regression}. Here, we need to define sample coalitions $z_j \in \{0, 1\}^p$, $j \in \{1, 2, \ldots, J\}$ where the value of 1 indicated the presence of corresponding variable in coalition, and the mapping function $h_x: \{0, 1\}^p \to \mathbb{R}^p$ that converts binary vectors into the original input space (1 represents the original value). In this setting, the Shapley kernel remains the same, and the weight given to each binary vector $z$ is
\begin{align}
w(z) = \frac{p - 1}{\binom{p}{s} s (p - s)},
\end{align}
where $s$ is the number of ones in $z$. Let $Z$ be the matrix of all binary vectors. Then SurvSHAP(t) is estimated as
\begin{align}
\Phi = (Z^TWZ)^{-1}Z^TWY,
\end{align}
where $W$ is the diagonal matrix consisting of Shapley kernel weights, and $Y$ is the matrix whose rows contain the survival function values predicted by the model $F$ for the mapping of each row of the $Z$ matrix by the $h_x$ function. Each row of the resulting $p \times r$ matrix $\Phi$ contains an explanation for a single variable included in the model.

Thanks to such algorithm structure, SurvSHAP(t) preserves the desired SHAP properties, stated in~\cite{shap} extended to consider the time-dependent nature of the explanation: local accuracy, missingness, and consistency.

In the context of this study, the property of local accuracy can be defined as:

\begin{align}
    \forall_t \; \hat{S}(t,\textbf{x}) = e_t^\emptyset + \sum_{d=1}^p \phi_t(\textbf{x}, d). \label{local-accuracy-property}
\end{align}

We use SurvSHAP(t) to calculate variable importance by aggregating the time-dependent function as 

\begin{align}
    \psi(\textbf{x}, d) = \int_0^{t_m} \left| \phi_{t}(\textbf{x}, d)\right| \, \mathrm{d}t.
\end{align}

\section{Evaluation metrics}

We introduce metrics used in Section \ref{sec:experiments} for assessing the quality of explanations.  

Local accuracy~\eqref{local-accuracy-metric} is a time-dependent adaptation of a local accuracy metric proposed in~\cite{treeshap}. It is calculated as the normalized standard deviation of the difference between the black-box model's output and the explanation as follows:

\begin{align}
    \sigma(t) = \sqrt{ \frac{\mathbb{E}(\hat{S}(t,\textbf{x})-\sum_{i}\phi_t(\textbf{x}, i))^2 } {\mathbb{E}\hat{S}(t,\textbf{x})^2 }}. \label{local-accuracy-metric}
\end{align}

Lower values of this metric indicate that for that specific time point, the sum of contributions of variables is closer to the actual output of the model. For methods that meet the local accuracy property, the values of this metric are zero. 

Another metric we propose to evaluate the ability of an explanation method to show the variables whose effect changes in time is Changing Sign Proportion (CSP). Its purpose is to numerically measure for what fraction of explained observations the value of SurvSHAP(t) was positive and negative for at least $\alpha$ of the considered time period. Specifically, it is defined as
{\small
\begin{align}
    CSP_{\alpha, t_{\mathrm{s}}, t_{\mathrm{e}}} = \frac{1}{n} \sum_{i=1}^n \mathlarger{\mathbbm{1}}\left(\left|\bigcup_{t\in[t_{\mathrm{s}},t_{\mathrm{e}}]}\{t:\phi_t(\textbf{x}_i, d)\geq0\}\right| > \alpha \cdot |[t_{\mathrm{s}},t_{\mathrm{e}}]| \right) \mathlarger{\mathbbm{1}}\left(\left|\bigcup_{t\in[t_{\mathrm{s}},t_{\mathrm{e}}]}\{t:\phi_t(\textbf{x}_i, d)\leq0\}\right| > \alpha \cdot |[t_{\mathrm{s}},t_{\mathrm{e}}]| \right), \label{CSP}
\end{align}
}%
where $n$ is the number of samples in the dataset, $t_{\mathrm{s}}$, $t_{\mathrm{e}}$ are start and end time points of selected time range,  $\phi_t(\textbf{x}_i, d)$ represents the Shapley value for observation $i$, variable $d$ and time $t$ and $|\cdot|$ is the length of the line segments. For variables with a time-dependent nature of an effect, the metric values should be greater than for variables whose type of effect is constant. In the case of models that do not take into account time-dependent effects, e.g., CPH, the values should be 0.

For evaluating the correlation between rankings of variables by their importance for prediction obtained from the explanation method and known ground-truth orderings (see Section \ref{sec:CPH-ranking}), we use the additive hyperbolic Kendall’s $\tau_h$ rank correlation coefficient \cite{weighted-tau}. We give the average of the coefficients obtained for all analyzed observations as a final measure.

In order to assess the SurvSHAP(t) quality against the ground-truth Shapley values, we use the GT-Shapley metric. It is adapted from \cite{treeshap} by applying it to each considered time point as follows:

\begin{align}
    \forall_t \; \rho(t) = \frac{1}{n}\sum_{i=1}^n\texttt{Pearson}([\phi_{t}(\textbf{x}_i, d)]_{1 \leq d \leq p}, [\phi^{true}_{t}(\textbf{x}_i, d)]_{1 \leq d \leq p}),  \label{gt-shapley}
\end{align}

where the values $\phi^{true}_{t}(\textbf{x}_i, d)$ are acquired using SurvSHAP(t) on a background sample with a much larger number of observations ($N=10 000$) generated from the same distribution.

Additionally, we define a metric based directly on residuals between obtained explanations and ground-truth values to measure how difficult it is for SurvSHAP(t) to explain a given variable $d$. It is given as follows:

\begin{align}
    \mathrm{normalized \: RMSE}(t, d) = \sqrt{\frac{\mathbb{E} (\phi_{t}(\textbf{x}, d) -  \phi^{true}_{t}(\textbf{x}, d))^2}{\mathbb{E} \phi^{true}_{t}(\textbf{x}, d)^2}} \label{normalized-rmse}.
\end{align}

\section{Experiments}
\label{sec:experiments}

Evaluation of machine learning explanations presents many challenges \cite{ml-clinical-dss, notions-of-explainability} as (1) in general, no ground truth of explanation is available \cite{liu2021synthetic}, and (2) one explains an imprecise black-box model provided imperfect data \cite{model-accuracy-explanation-quality}. Thus, for a comprehensive evaluation scheme, we divide our experiments into three steps:
\begin{enumerate}
    \item Measuring local accuracy and time-dependence on synthetic data
    \item Comparison with SurvLIME to show that SurvSHAP(t) is able to detect variables with a time-dependent effect, and its aggregation is a better determinant of the importance of variables for a prediction than SurvLIME.
    \item Showing in a real-world use case that SurvSHAP(t) properly explains machine learning survival models predicting heart failure.
\end{enumerate}

\subsection{Evaluating explanations on synthetic data}

\paragraph{Setup.} For the first experiment, data is generated synthetically in order to demonstrate that SurvSHAP(t) explanations work correctly for variables that have a time-dependent effect, provided that the used model can make use of such dependencies. The dataset \texttt{EXP1} consisting of $N=1000$ observations is generated using the method suggested for generating time-dependent effects by~\citet{simulating-data}. The base hazard function is defined as

\begin{align}
    h_0(t) = \exp(-17.8 + 6.5t - 11\sqrt{t}\cdot \ln{t} + 9.5 \sqrt{t}),
\end{align}

and for a chosen observation from the dataset, the hazard function is of the form

\begin{align}
    h(t) = h_0(t) \cdot \exp[ (-0.9 +0.1t +0.9\ln(t) ) x^{(1)} + 0.5 x^{(2)} - 0.2  x^{(3)}+ 0.1 x^{(4)} +10^{-6} x^{(5)}]. \label{generated-hazard}
\end{align}

The coefficients were chosen such that the variable $x^{(1)}$ has a time-dependent effect, $x^{(2)}, x^{(3)}$, and $x^{(4)}$ are variables with constant effect and $x^{(5)}$ is insignificant -- represents random noise. Variables $x^{(1)}$ and $x^{(2)}$ are binary, sampled from the binomial distribution, such that $\mathbb{P}(x^{(1)} = 0) = \mathbb{P}(x^{(1)} = 1) = \mathbb{P}(x^{(2)} = 1) = \mathbb{P}(x^{(2)} = 0) = 0.5$, whereas $x^{(3)} \sim \mathcal{N}(10, 2)$, $x^{(4)} \sim  \mathcal{N}(20, 4)$, and $x^{(5)} \sim  \mathcal{N}(0, 1)$.

To generate the survival times $T_i$, the process described in~\cite{simulating-data} is followed. In the first step, the hazard function~\eqref{generated-hazard} is integrated numerically to obtain the cumulative hazard function, which is then transformed into the survival function using formula~\eqref{relationship-S-H}. Then Brent's iterative root finding method is applied to function $g(t) = S(t,\textbf{x}) - U$, where $U \sim U[0,1]$. The found root is used as the true (latent) survival time $T_i$ for a vector of variables~$\textbf{x}_i$.

In order to determine observed times $y_i$ based on generated survival times $T_i$, a method proposed in~\cite{simulating-censoring} is used. For each observation, two values are generated from the uniform distribution: $C_{l,i} \sim U[11,16]$, which can be interpreted as the time to administrative censoring event, and $C_{r,i} \sim U[0,24]$ which denotes the time to the occurrence of a right censoring event. If both those values are higher than the generated time $T_i$ then $\delta_i=1$, otherwise $\delta_i=0$. The observed time $y_i$ is defined as $y_i = \min\{T_i, C_{ri}, C_{li}\}$, which in this case translates into a censoring rate of 0.331.

\begin{wrapfigure}[19]{r}{0.4\textwidth}
  \centering
  \includegraphics[width=0.4\textwidth]{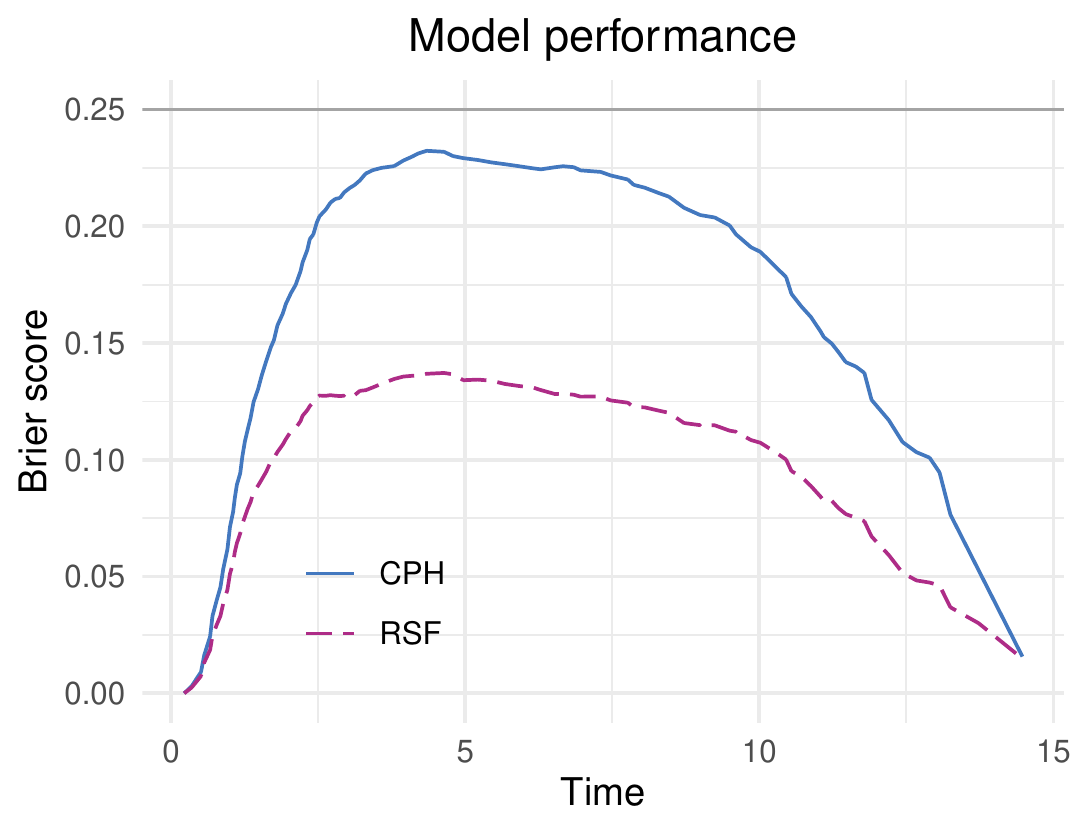}
  \caption{Time-dependent performance of the RSF and CPH models measured by Brier score (lower is better; Brier score of 0.25 indicates random predictions).}
  \label{fig:exp1_brier_score}
\end{wrapfigure}
\newpage
\paragraph{Results. } The first experiment intends to show how the SurvSHAP(t) works. Thus, we fit Cox Proportional Hazards and Random Survival Forest models to the generated dataset and calculate the prediction explanations for each observation. The models' performance expressed in the Brier score measure \cite{graf} is presented in Figure \ref{fig:exp1_brier_score}. RSF has an integrated Brier score equal to 0.097 and, because of its ability to model complex dependencies, outperforms CPH, for which the integrated Brier score is 0.167.

\begin{figure}[b!]
  \centering
  \includegraphics[width=\textwidth]{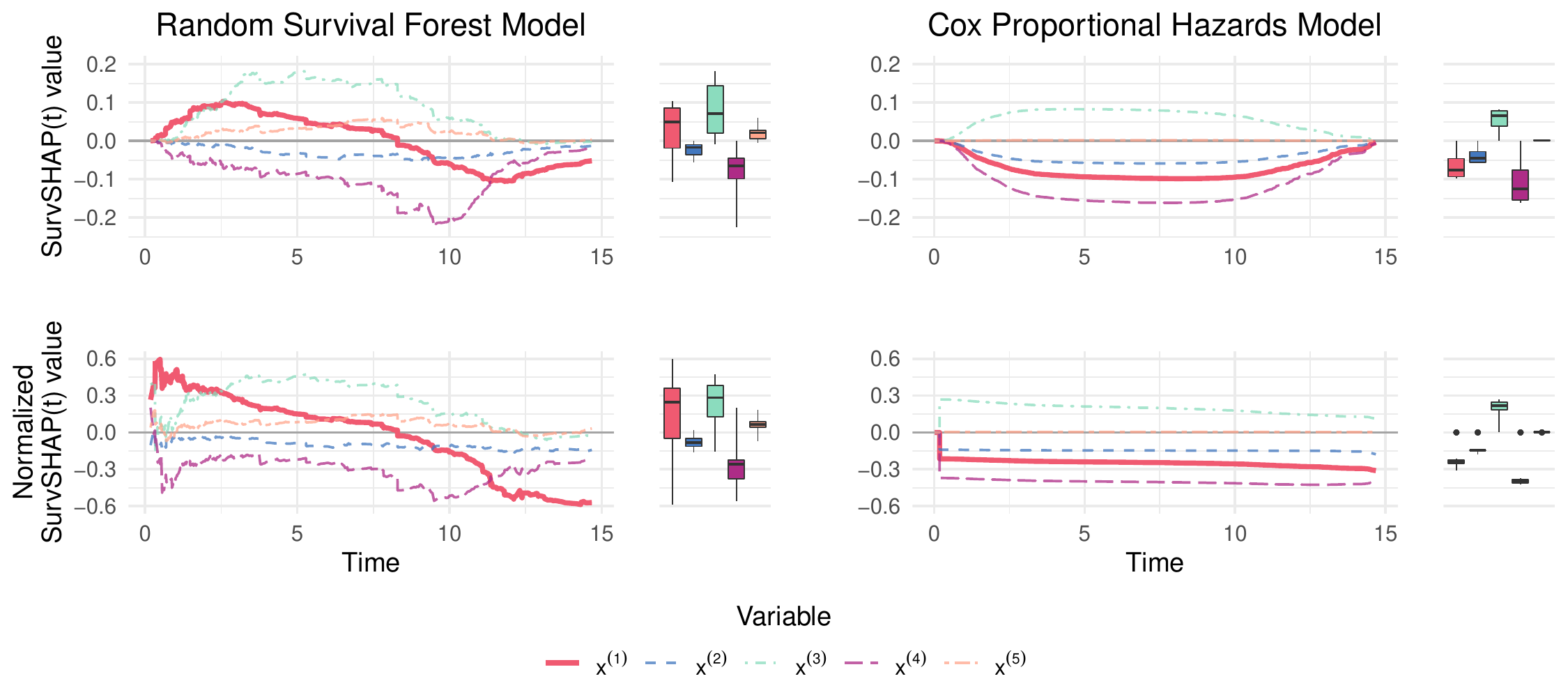}
  \caption{SurvSHAP(t) for the selected observation and two models trained on the dataset \texttt{EXP1}. }
  \label{fig:exp1_example_shap}
\end{figure}

In Figure~\ref{fig:exp1_example_shap} the SurvSHAP(t) functions for a selected prediction are presented. In the first row, SurvSHAP(t) functions of each variable are shown for each of the two models, whereas in the second row, they are normalized according to formula~\eqref{SHAP-normalization}. Positive SurvSHAP(t) values indicate that a given variable has increased the survival function by that much, while negative values indicate a decrease. It can be seen that the variable $x^{(1)}$, which has a time-dependent effect (positive at the beginning, negative later), is correctly modeled in RSF but not in CPH. This shows that SurvSHAP(t) is capable of finding such differences between models (i.e., it explains the model, not data) and therefore is useful for validating if models consider time-dependent variables. Indeed, by looking at the normalized SurvSHAP(t) values, we can see that the variable effects for the CPH are constant over time (narrow boxplots). Moreover, RSF assigned part of the changing impact over time to other variables -- it is expected as a non-parametric model without knowledge of a specific form of the time-dependent effect has difficulties with its precise separation (i.e., determination of the source of this effect).

Another benchmark we perform is checking if the additivity property of SHAP is retained. For this purpose, we computed the time-dependent version of the local accuracy metric defined in \eqref{local-accuracy-metric}.

\begin{wrapfigure}[25]{R}{0.4\textwidth}
  \centering
  \includegraphics[width=0.4\textwidth]{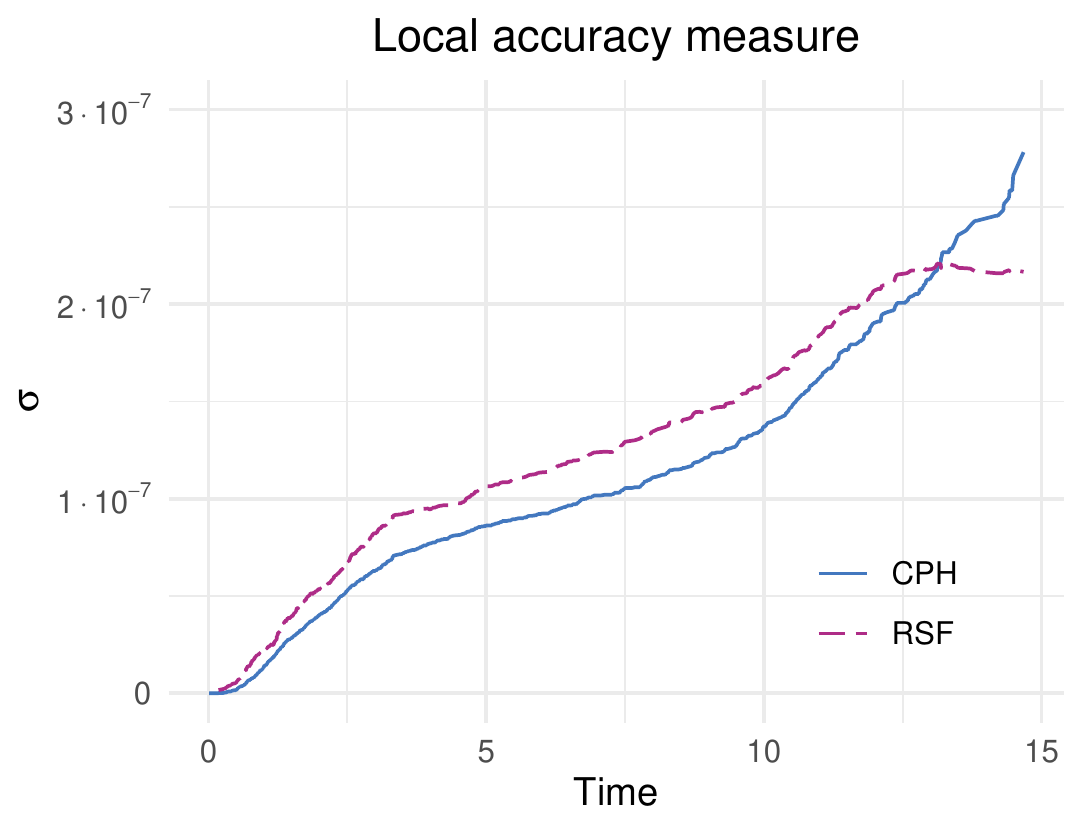}
  \caption{Sanity check of local accuracy (additivity) property for two models trained on the dataset \texttt{EXP1}.}
  \label{fig:exp1_local_accuracy}
\end{wrapfigure}

Low values of the local accuracy metric (shown in Figure \ref{fig:exp1_local_accuracy}) on the order of $10^{-7}$ indicate that the property is preserved. The fact that they seem to rise with time is also expected as it is difficult for models to make good predictions near the end of the examined time, where many observations are censored.

As the data used for this experiment is synthetically generated, we know that the variable $x^{(1)}$ has a time-dependent effect, positive at the beginning and negative later. Therefore, if a model uses this fact in its prediction, it should be noticeable in the explanations, such that in some proportion of the considered time period, the effect is negative and in some -- positive. We use the Changing Signs Proportion metric \eqref{CSP} defined earlier with start and end time points fixed at 0.1 and 0.9 quantiles of the times included in the data, respectively.

\newpage

The value should be low for variables with a constant effect on the prediction and noticeably higher for the variables whose effect changes in time. It should be very low for models we know cannot take time-dependent variables into account, e.g., the Cox Proportional Hazards model. Table \ref{tab:exp1_changing_signs} presents the values of $CSP_{0.05}$ for each variable and for both models. We see that the metric is high for the $x^{(1)}$ variable for RSF and low for other variables. For CPH, the metric is close to 0, as variables can only have either positive or negative influence \emph{independent of time}.

\newlength{\oldintextsep}
\setlength{\oldintextsep}{\intextsep}
\setlength\intextsep{-10pt}

\begin{wraptable}[10]{l}{0.57\textwidth}
\caption{Comparison of the $CSP_{0.05}$ metric values for each variable between Random Survival Forest and Cox Model.}
\vspace{1em}
\centering
\begin{tabular}{crr}
\toprule
\textbf{Variable} & \multicolumn{1}{c}{\textbf{RSF}} & \multicolumn{1}{c}{\textbf{CPH}}\\
\midrule 
$x^{(1)}$                & 0.954                            & 0.000                            \\
$x^{(2)}$                & 0.127                            & 0.000                            \\
$x^{(3)}$                & 0.154                            & 0.072                            \\
$x^{(4)}$                & 0.274                            & 0.066                            \\
$x^{(5)}$                & 0.481                            & 0.015                            \\ 
\bottomrule
\end{tabular}
\vspace{2em}
\label{tab:exp1_changing_signs}
\end{wraptable}

\setlength\intextsep{\oldintextsep}
\begin{wrapfigure}[18]{r}{0.4\textwidth}
  \includegraphics[width=0.4\textwidth]{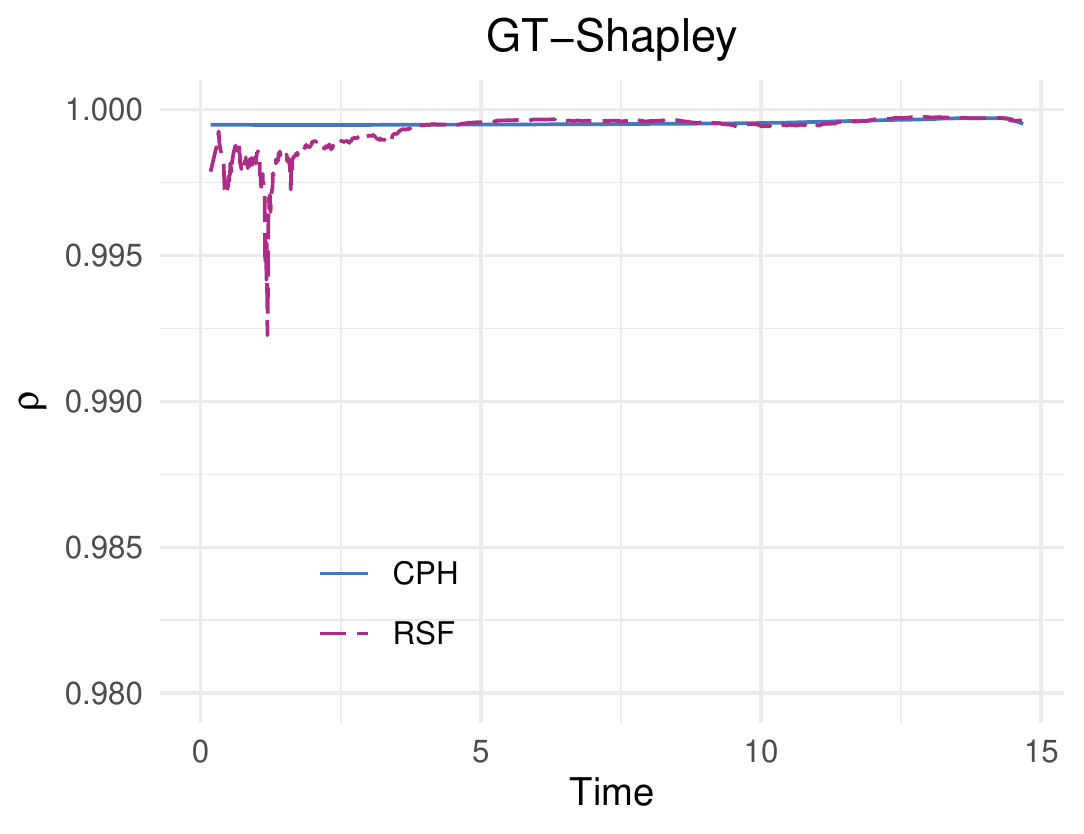}
  \caption{Comparison of the GT-Shapley metric for explanations of CPH and RSF.}
  \label{fig:exp1_corr}
\end{wrapfigure}

Moreover, we measured the value of the GT-Shapley metric \eqref{gt-shapley} for explanations of CPH and RSF models. For each time point, we calculate the Pearson's correlation between SurvSHAP(t) generated on the basis of the \texttt{EXP1} dataset and a larger, artificially generated sample of observations (as we know the underlying distribution of data).  The high scores shown in Figure~\ref{fig:exp1_corr} confirm that the explanations are stable: a greater number of points to estimate SurvSHAP(t) does not meaningfully change the explanations. 

\begin{figure}[b!]
  \centering
  \includegraphics[width=0.89\textwidth]{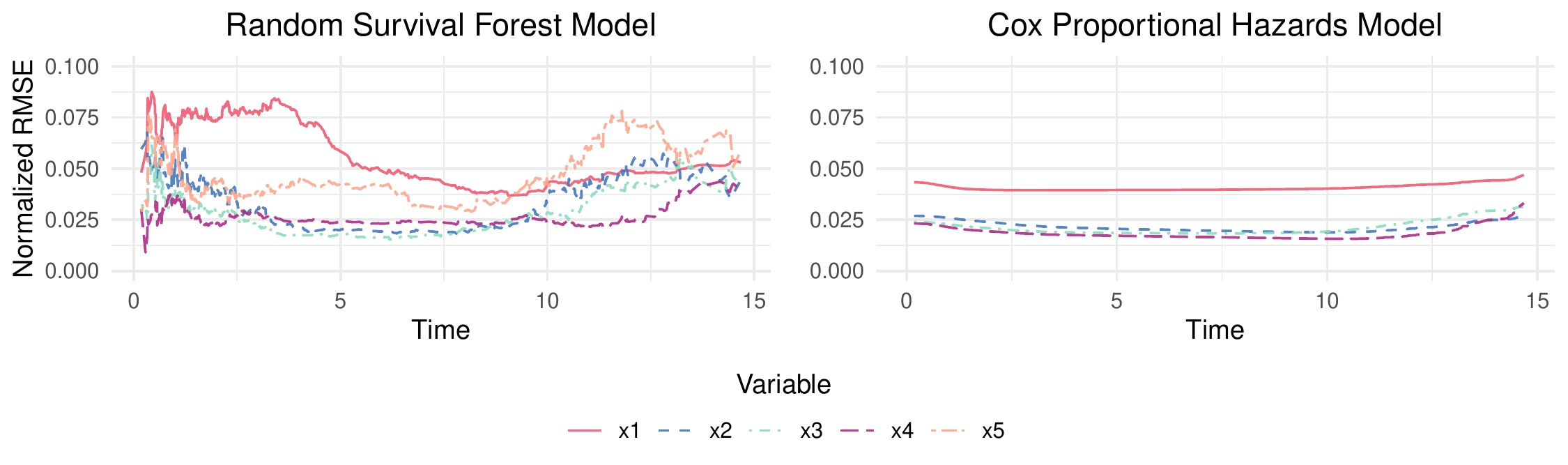}
  \caption{Comparison of normalized RMSE of SurvSHAP(t) for each variable between RSF and CPH (\textbf{lower} is better). \textbf{Note:} variable $x^{(5)}$  is off the scale for CPH.}
  \label{fig:exp1_error}
\end{figure}

We also evaluated the explanations of CPH and RSF models using the normalized RMSE metric \eqref{normalized-rmse}, which is visualized in Figure~\ref{fig:exp1_error}. It is clearly visible that for both models, the attribution of the time-dependent variable $x^{(1)}$ is difficult to explain. We suspect that time-dependent variables need a bigger background of observations to be explained correctly. Another fact worth noting is that explanations for the variable $x^{(5)}$ perform much worse in the CPH model -- the plot needs to be cropped to show useful information. For $x^{(5)}$, the attributed SurvSHAP(t) values are close to 0, as we see in Figure~\ref{fig:exp1_example_shap}, so a reason for the high value of the metric might be the numerical errors occurring when normalizing such low values.

\subsection{Comparison to SurvLIME}
\label{sec:compare-survlime}

\paragraph{Setup.} We aim to compare the explanations of SurvSHAP(t) with those provided by SurvLIME and therefore follow the same experimental setup as in~\cite{survlime}. We use two datasets where the vector of variables $\textbf{x}_i\in \mathbb{R}^5$ is generated from the uniform distribution on a 5-dimensional sphere with a predefined radius. The center of the sphere in \texttt{dataset0} is $(0,0,0,0,0)$, whereas \texttt{dataset1} is sampled from a sphere centered around $(4, -8, 2, 4, 2)$. Both spheres have the radius $R=8$. The survival times for these data are generated according to the method proposed in~\cite{generating-weibull} with the following formula:
\begin{align}
    y_i = \left( \frac{-\ln{U}}{\lambda \exp(\textbf{b}^T\textbf{x}_i)}\right)^{1/v},
\end{align}
where $U \sim U[0,1]$ and
\begin{itemize}
    \item in \texttt{dataset0}: $\lambda = 10^{-5}$, $v = 2$, $\textbf{b}^T = (10^{-6}, 0.1, -0.15, 10^{-6}, 10^{-6})$,
    \item in \texttt{dataset1}: $\lambda = 10^{-5}$, $v = 2$, $\textbf{b}^T = (10^{-6}, -0.15, 10^{-6}, 10^{-6}, -0.1)$.
\end{itemize}
Event indicators $\delta_i$ are generated from the binomial distribution, such that $\mathbb{P}(\delta_i = 1) = 0.9$ and $\mathbb{P}(\delta_i = 0) = 0.1$. Each dataset consists of $N=1000$ observations divided into train and test sets in the 9:1 proportion. We use the test sets to train the model and then evaluate it and explain it in the test setting.

\begin{wrapfigure}[27]{l}{0.4\textwidth}
   \includegraphics[width=0.4\textwidth]{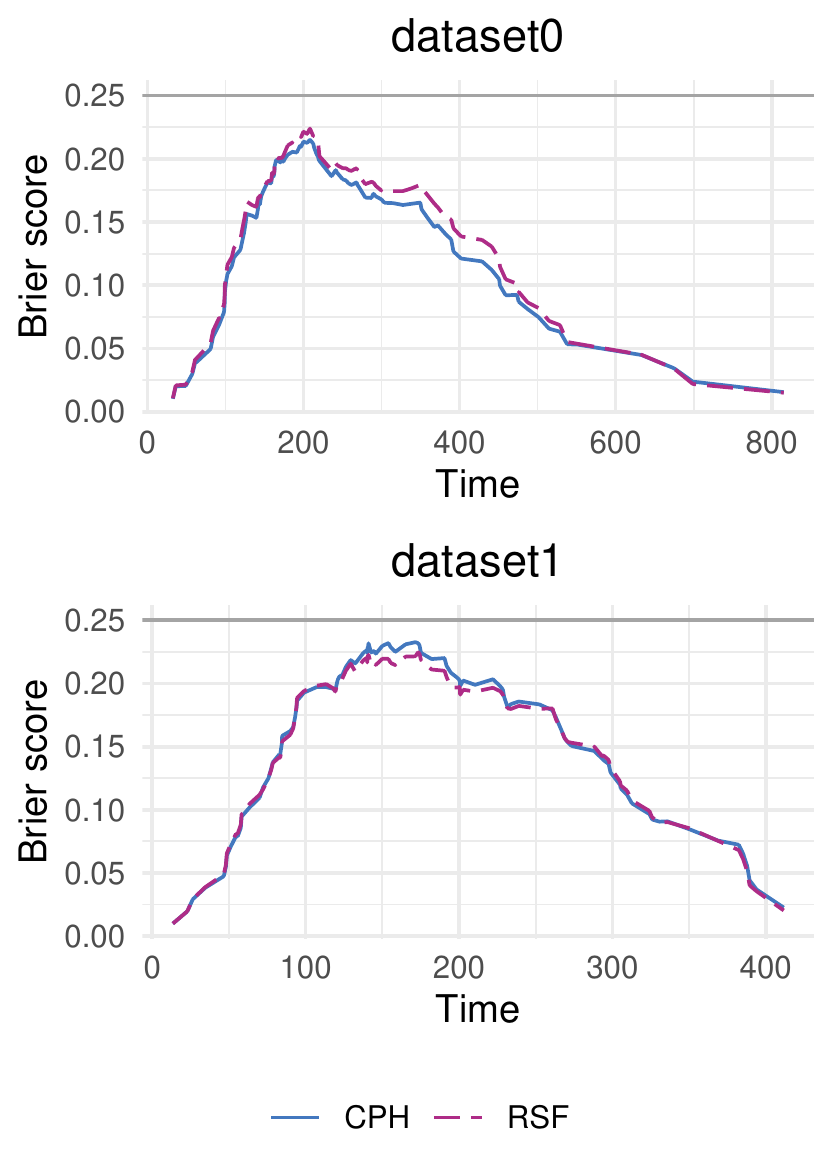}
  \caption{Time-dependent performance of the RSF and CPH models measured by Brier score (lower is better; Brier score of 0.25 indicates random predictions).}
  \label{fig:exp2_brier_score}
\end{wrapfigure}

\begin{wrapfigure}[13]{r}{0.4\textwidth}
  \centering
  \includegraphics[width=0.4\textwidth]{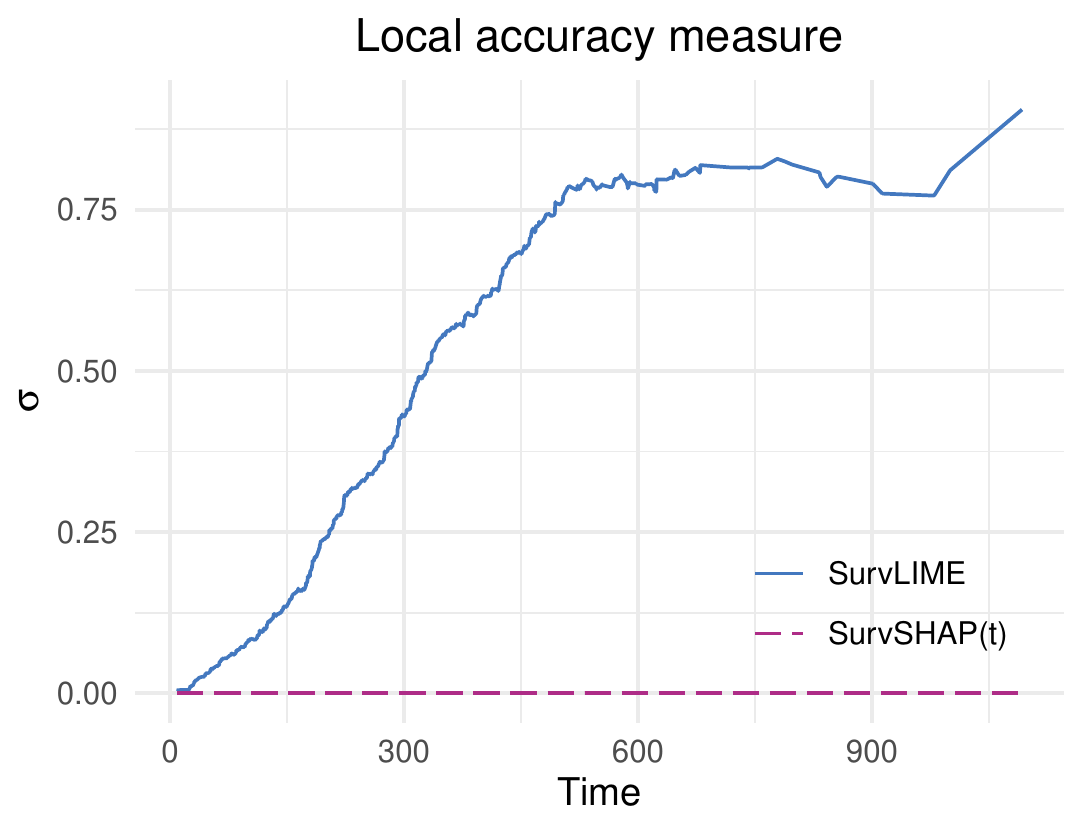}
  \caption{Normalized standard deviation of the difference between black-box model output and the explanation (lower is better). \textbf{Note:} the curve for SurvSHAP(t) coincides with the x-axis.}
  \label{fig:exp2_local_accuracy_rsf_cluster_0}
\end{wrapfigure}

\paragraph{Results.} For this experiment, four models were fitted, i.e., the Cox Proportional Hazards and Random Survival Forest models for each of the two datasets. For \texttt{dataset0} the values of integrated Brier score indicating the performance of the models were 0.103 for RSF and 0.097 for CPH, and for \texttt{dataset1} RSF achieved a score of 0.137, whereas CPH 0.139. The values of the Brier score for the entire considered time range are presented in Figure~\ref{fig:exp2_brier_score}. The first performed test confirms that SurvSHAP(t) preserves the local accuracy property, whereas SurvLIME does not. The values of normalized standard deviations of local accuracy \eqref{local-accuracy-metric} for RSF model trained on \texttt{dataset0} are presented in Figure~\ref{fig:exp2_local_accuracy_rsf_cluster_0}. We see that the value for SurvSHAP(t) is close to 0 across the whole time range, while for SurvLIME, it is significantly larger. This is expected as SurvSHAP(t) can explain all functions, whereas the explanation of SurvLIME always takes the form of a Cox model's survival function.

Both SurvLIME and SurvSHAP(t) can be used to assess the relative importance of variables on the local level (i.e., for selected prediction). First, we fit Cox Proportional Hazards model to obtain the ground-truth variable importance ranking via the method described in Section~\ref{sec:CPH-ranking} to both datasets. We calculate the rankings of variable attributions using SurvLIME and SurvSHAP(t) for each observation from the test set. Then we use the additive hyperbolic Kendall's $\tau_h$ rank correlation coefficient to determine the similarity between the true and explanation rankings. Finally, we average this value across all observations. Table~\ref{tab:exp2_kendall-tau} presents the results proving better variable importance ranking estimation of SurvSHAP(t) in the glass-box evaluation scheme.

\begin{table}[b!]
    \caption{Average $\tau_h$ correlations of the variable importance rankings according to explanations against the ground-truth ranking in the Cox model (\textbf{higher} is better).}
    \vspace{1em}
    \label{tab:exp2_kendall-tau}
    \centering
    \begin{tabular}{crr}
    \toprule
     & \multicolumn{1}{c}{\textbf{SurvLIME}} & \multicolumn{1}{c}{\textbf{SurvSHAP(t)}} \\ 
    \midrule
    \texttt{dataset0} & 0.763 & 0.917 \\
    \texttt{dataset1} & 0.454 & 0.745 \\          
    \bottomrule                  
    \end{tabular}
\end{table}

\newpage
Consecutively, we fit a black-box Random Survival Forest to both datasets and explain its predictions using both methods. The ground-truth importance ranking of the RSF black-box model is not available.

\begin{figure}[b!]
  \centering
  \includegraphics[width=\textwidth]{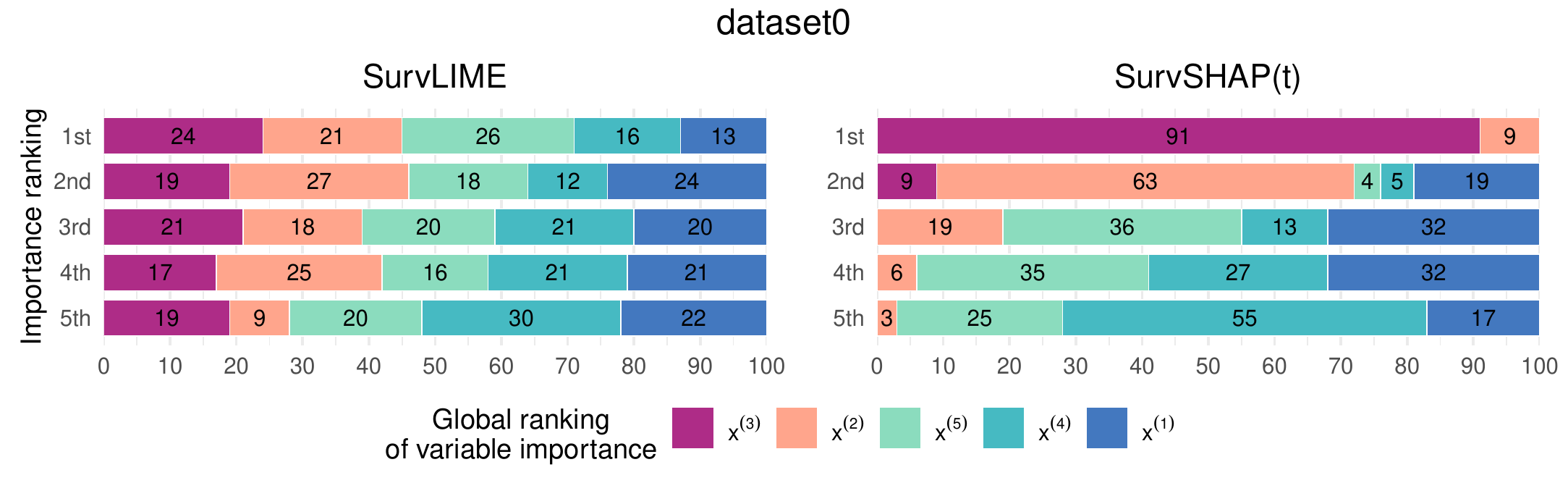}
  \includegraphics[width=\textwidth]{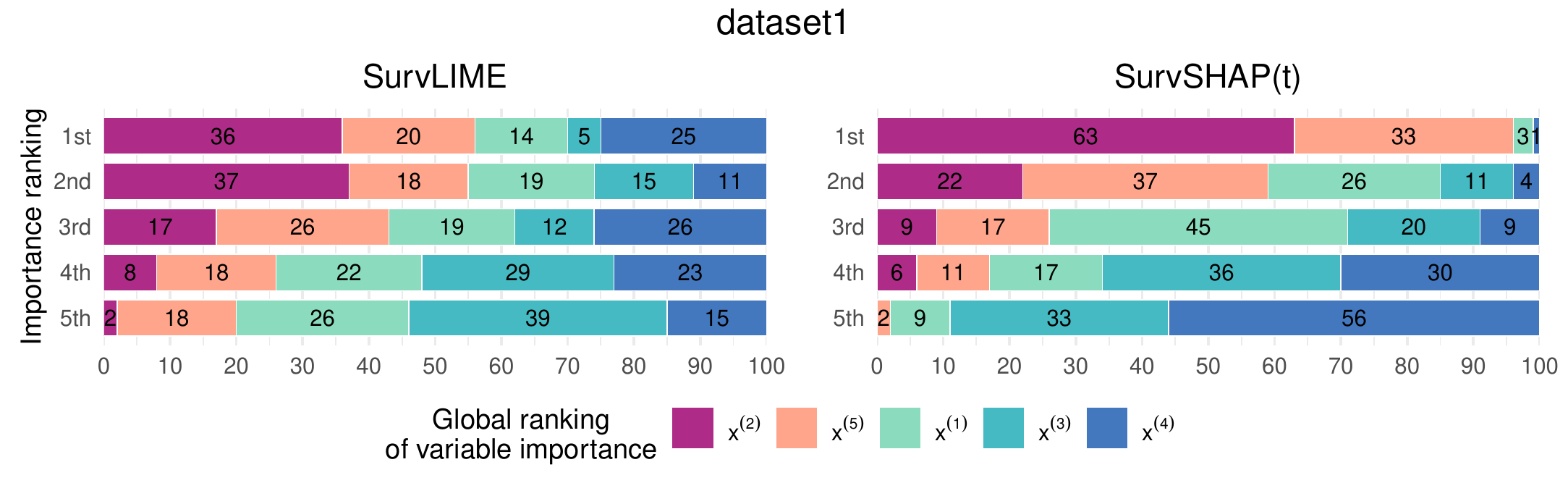}
  \caption{Juxtaposition of local and global importance rankings for 100 predictions of the RSF model fitted to~\texttt{dataset0} (\textbf{top}) and \texttt{dataset1} (\textbf{bottom}). In each case, there are two globally important variables and three less important ones--the colors are specifically sorted to show the global ranking of variables. We observe that SurvSHAP(t) maintains the majority observations per each consecutive variable (specifically in \texttt{dataset1} the majorities are represented by 63 for $x^{(2)}$ in 1st, 37 for $x^{(2)}$ in 2nd, 45 for $x^{(1)}$ in 3rd, 36 for $x^{(3)}$ in 4th, and 56 for $x^{(4)}$ in 5th) outperforming SurvLIME, which provides more uniformly distributed rankings (especially in \texttt{dataset0}). }
  \label{fig:exp2_rsf_orderings}
\end{figure}

Therefore, we imitate the ranking using permutational variable importance \cite{permutational-vi} with the integrated Brier score as a loss function. Figure~\ref{fig:exp2_rsf_orderings} visualizes the aggregation of variable rankings over 100 observations in the test set. Each horizontal bar represents the fraction of observations for which the variable represented as a given color was ranked 1st, 2nd, etc. The correct ordering, obtained by permutational variable importance, is presented in the legend of the Figure. We observe that the global aggregation of local SurvLIME rankings is close to random -- for \texttt{dataset0} almost every place has a uniform distribution of variables, whereas SurvSHAP(t) distinctly attributes one variable to the first and second place in the ranking.

\newpage

\begin{wrapfigure}[20]{r}{0.4\textwidth}
  \centering
  \includegraphics[width=0.4\textwidth]{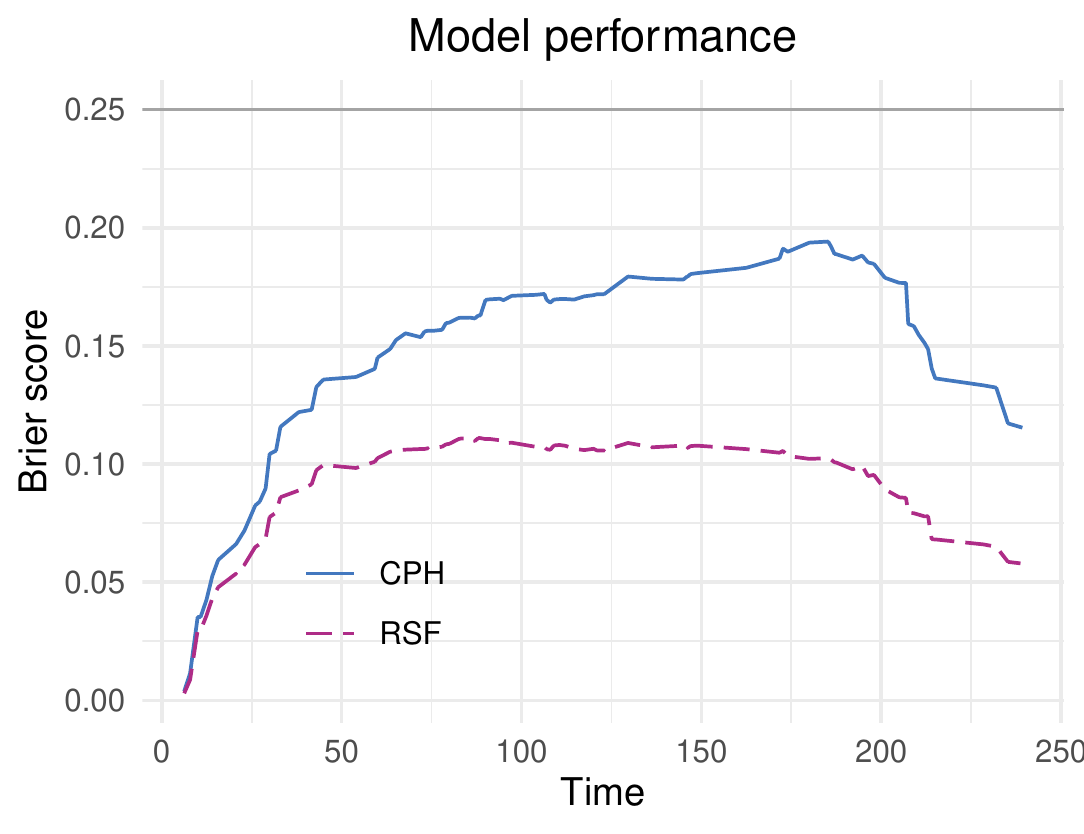}
  \caption{Time-dependent performance of the RSF and CPH models measured by Brier score (lower is better; Brier score of 0.25 indicates random predictions).}
  \label{fig:exp3_brier_score}
\end{wrapfigure}

\subsection{Real-world use case: predicting survival of patients with heart failure}

The primary motivation for the development of SurvSHAP(t) is the potential of such a method in practical applications. In analyses of medical time-to-event data, time-dependent effects, i.e., non-proportional hazards, often occur \cite{donizy2014nuclear, jatoi2011breast, mok2009gefitinib}. The proposed method can confirm that the model of interest captures such dependencies. However, SurvSHAP(t) gives valuable insights into the operation of the model not only when such a phenomenon occurs. Provided explanations may help to interpret and understand critical decisions or generate new domain knowledge.

In order to show the use case of the SurvSHAP(t), we apply the method to two models trained on the real-world \texttt{heart\_failure} dataset. The analyzed cohort includes records of 299 heart failure patients collected at the Faisalabad Institute of Cardiology and the Allied Hospital in Faisalabad (Punjab, Pakistan) \cite{ml-heart-failure}. We use eight  variables selected based on previous results (the most important according to accuracy decrease). In this case, we also use two algorithms: Cox Proportional Hazards and Random Survival Forest models. The performance of the models measured with the Brier score is presented in Figure~\ref{fig:exp3_brier_score}. The integrated Brier score for the RSF model is 0.093, while the CPH model has a score of 0.152.

\begin{figure}[b!]
  \centering
  \includegraphics[width=0.97\textwidth]{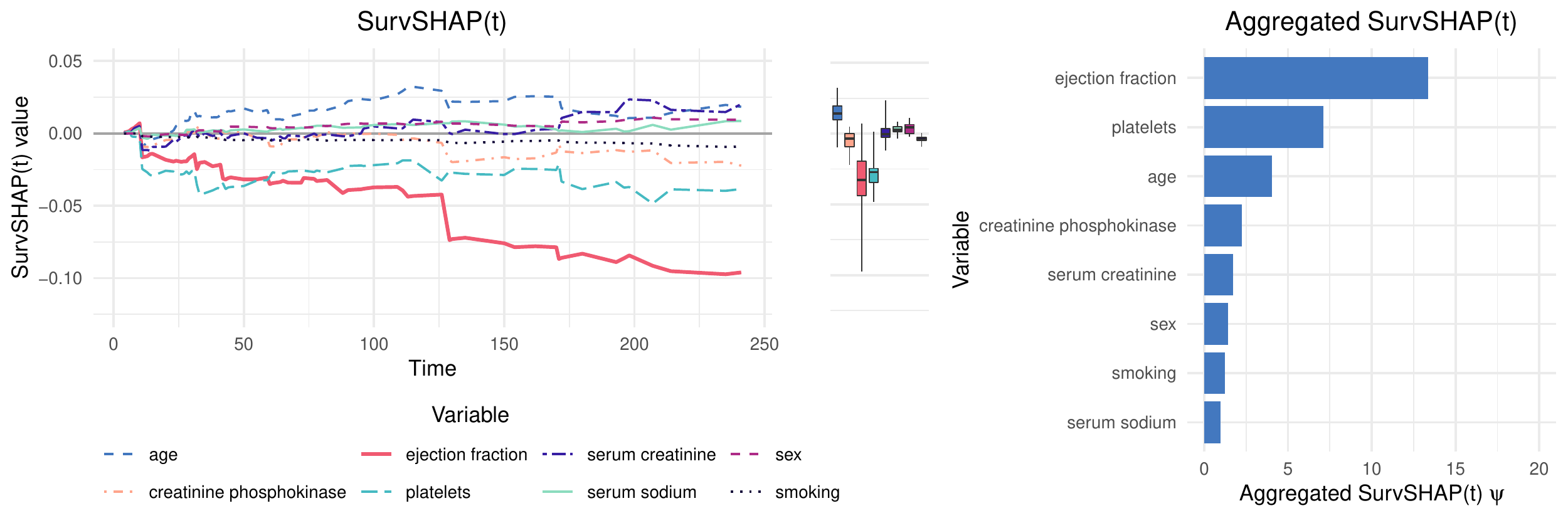}
  \caption{Explanation results for the selected observation and RSF model trained on the \texttt{heart\_failure} dataset: SurvSHAP(t) values (\textbf{left}) and aggregated SurvSHAP(t) values -- variable importance measure (\textbf{right}).}
  \label{fig:exp3_intro_shap}
\end{figure}

\begin{figure}[b!]
  \centering
  \includegraphics[width=0.97\textwidth]{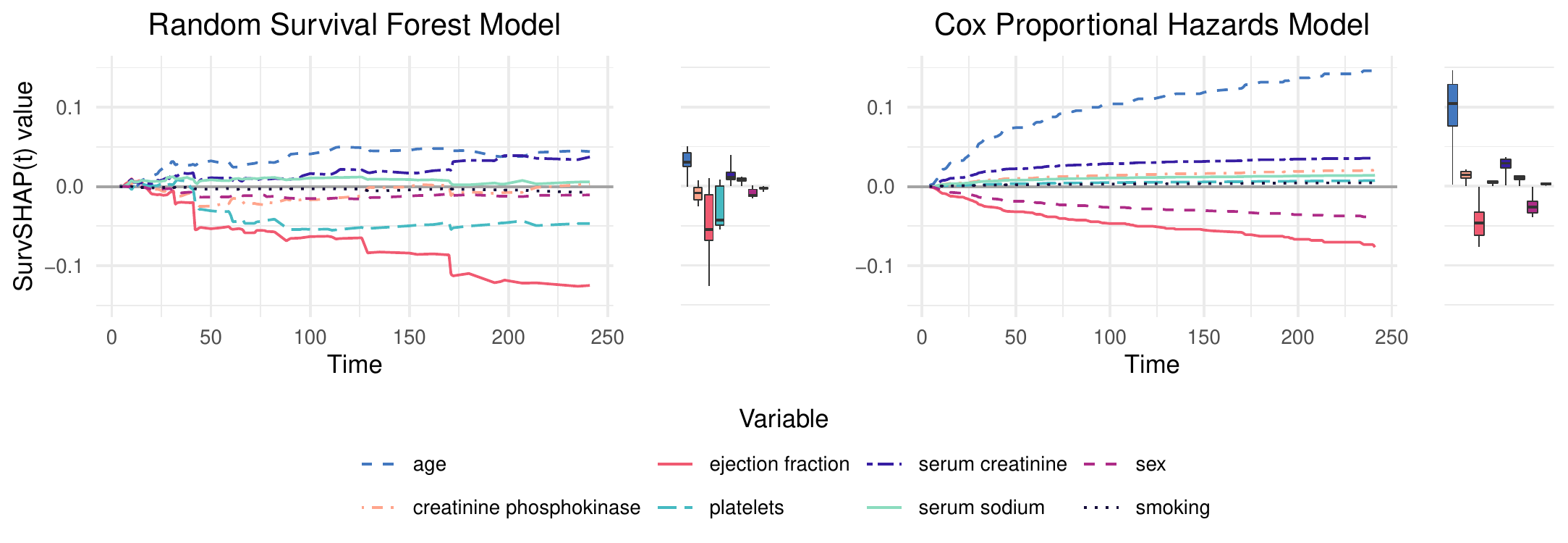}
  \caption{SurvSHAP(t) for the selected observation and two models trained on the \texttt{heart\_failure} dataset.}
  \label{fig:exp3_example_shap_2}
\end{figure}

An exemplary single prediction explanation for the RSF model is presented in Figure~\ref{fig:exp3_intro_shap}. It shows that the importance of ejection fraction for the model increases over time which is indicated by the decrease in SurvSHAP(t) value for this particular variable.  We also see which other variables impact the obtained prediction most. With domain knowledge, it allows for assessing whether a given model output is reliable.

In Figure~\ref{fig:exp3_example_shap_2}, explanations of the decisions of both models for the same patient are presented. They show some inconsistencies between the models (e.g., for the \texttt{platelets} variable). In-depth analysis enables the physician to decide which prediction is more adequate, helping to develop personalized medicine.

It is challenging to obtain ground-truth Shapley values for such a small sample and real complex data. Moreover, due to the presence of binary variables in the data, it becomes impossible to use the importance of variables in the Cox model as defined in Section~\ref{sec:CPH-ranking}. Therefore, we imitate both models' global variable importance ranking by calculating the permutational variable importance again. Further, we compare the importance rankings of the variables obtained by aggregating SurvSHAP(t) with those from the Cox model found by SurvLIME.

\begin{figure}[b!]
  \centering
  \includegraphics[width=\textwidth]{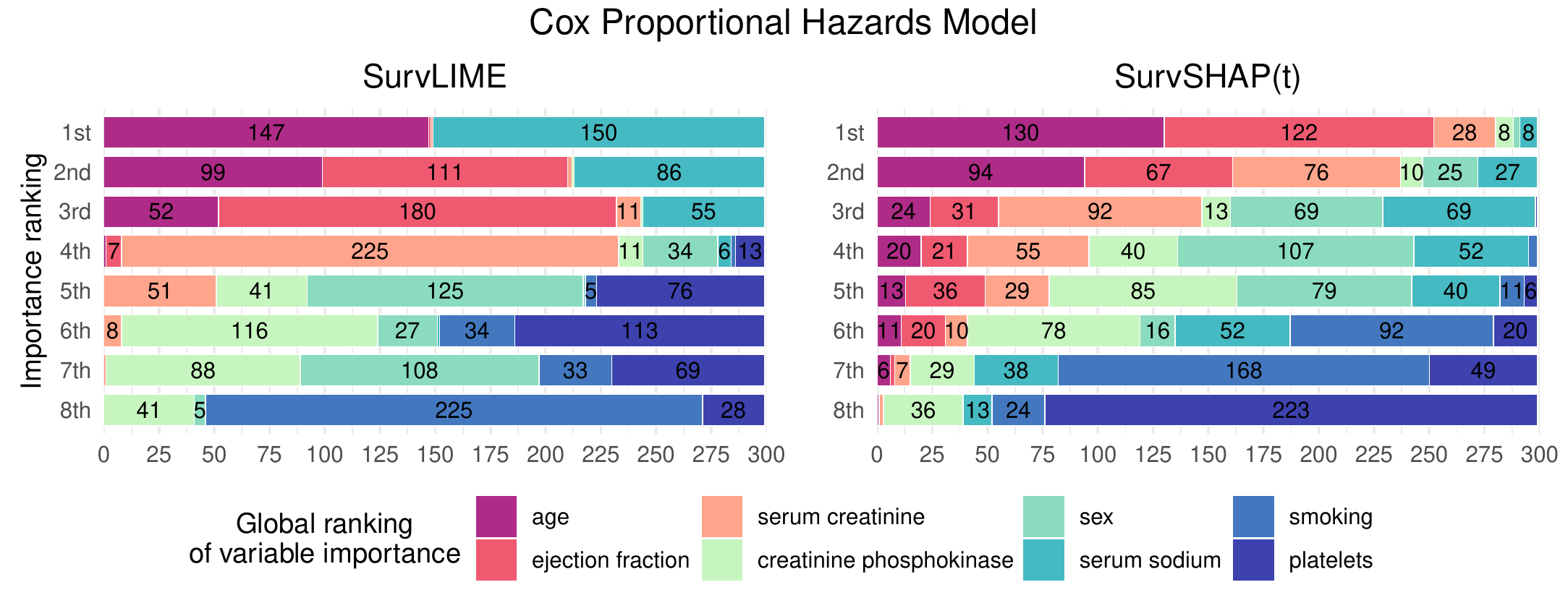}
  \includegraphics[width=\textwidth]{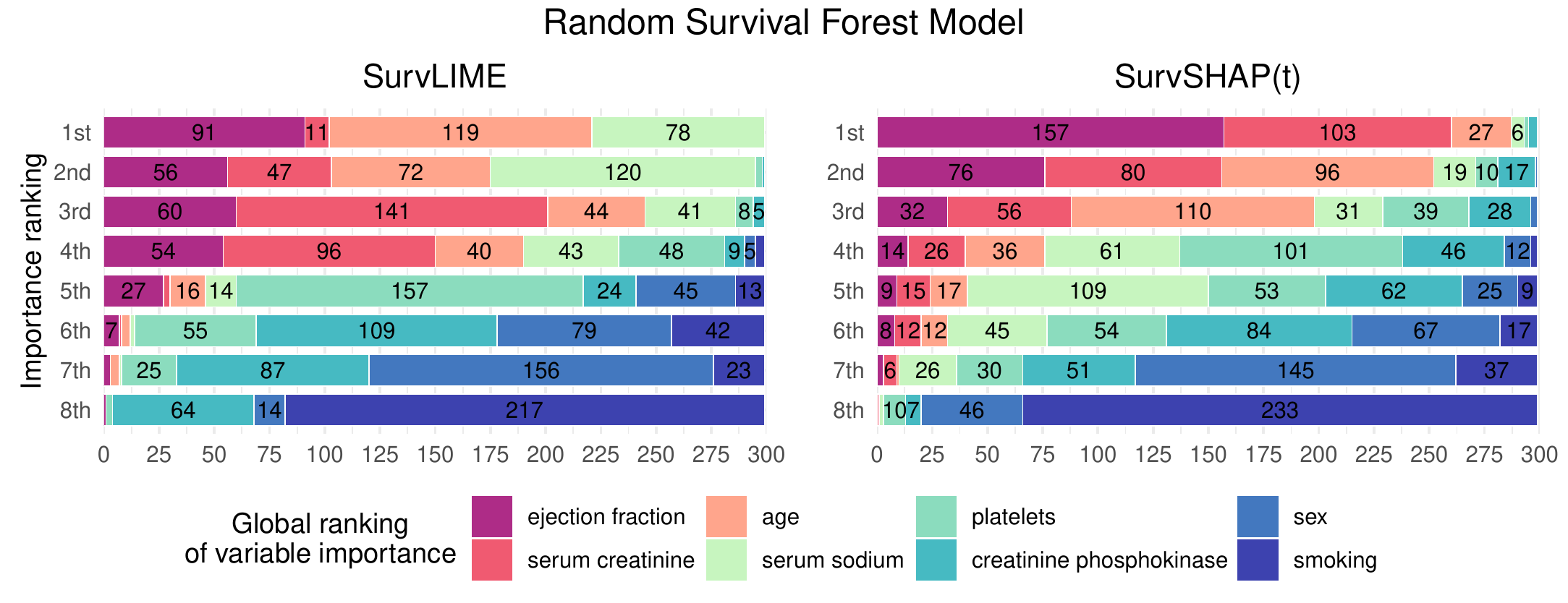}
  \caption{Juxtaposition of local and global importance rankings for predictions of the CPH model (\textbf{top}) and RSF (\textbf{bottom}). The colors are specifically sorted from purple to blue showing the global ranking of variables in each model.}
  \label{fig:exp3_orderings}
\end{figure}

The results are presented in Figure~\ref{fig:exp3_orderings}, where the variables within one bar are sorted by the global importance. Again, one can observe the superiority of SurvSHAP(t) aggregation over the coefficients derived from the SurvLIME method. For the Cox model, SurvLIME most often indicates the \texttt{serum sodium} as the most important variable, the third least important variable globally. SurvSHAP(t) indicates \texttt{age} as the most important variable for the biggest percentage of the CPH predictions, which is consistent with the global ranking. Our method coped even better in the context of assessing the importance of variables in the RSF model. It identified \texttt{ejection fraction} and \texttt{serum creatinine} as the most important variables 260 times in total. These two variables have comparable importance globally (the mean increase of the integrated Brier score after a permutation is 0.0456 and 0.0454, respectively).

\newpage
\section{Discussion}

SurvSHAP(t) extends the idea of SHAP to a broad class of models working on survival data. It is a subsequent method designed to explain survival functions, but the first one based on the concept of Shapley values and the first that provides time-dependent explanations. 

\paragraph{On the differences to SurvLIME.} SHAP has been chosen as the basis of this method because it is one of the most widely used approaches for explaining black-box models~\cite{holzinger2022xaioverview}. It was also suggested as an area of possible further research in \cite{survlime}. It is important to notice that the approach to explanation is different from the one proposed in SurvLIME. For the calculation, the entire output function is considered in both methods, but SurvLIME finds the closest possible CHF among the outputs of a Cox model, whereas SurvSHAP(t) does not have this limitation. The authors of SurvLIME state that the independence of the Cox model's variable effect on time is an advantage, as it makes the optimization problem significantly simpler. However, it has one drawback -- SurvLIME cannot properly explain variables whose effect changes in time. SurvSHAP(t) does not make use of the Cox model, so it is able to explain variables that have a time-dependent effect. The change is also visible in the context of the final explanation form. SurvLIME outputs the found coefficients of a Cox model (each variable's attribution described by a single value), while SurvSHAP(t) produces functions of time-dependent importance for each variable. Moreover, coefficients of the SurvLIME explanation do not directly indicate the importance of variables, i.e., the magnitude of the variable is also a contributing factor. The proposed method presents importance at all time points explicitly.

\paragraph{Software implementation.} The inclusion of code implementing both SurvSHAP(t) and SurvLIME in Python is another key point of the conducted research, as it allows for the application of explanations to existing models. The code is tailored to work with the models implemented in the \texttt{scikit-survival} Python package \cite{sksurv}. The produced plots give the user an intuitive visualization of the SurvSHAP(t) explanations -- one can see what the influence of a particular variable is at any chosen time, even if they do not have previous experience with XAI.

\paragraph{Limitations.} Another thing worth noting is the fact that SurvSHAP(t), as an extension of SHAP, inherits many of its drawbacks. One of them is that reported values might be misleading if the model is not additive~\cite{ema}. Another practical limitation is the fact that the computation of Shapley values is time-consuming, which is amplified even more by the fact that the calculation needs to be done for many time points.

\section{Conclusion}

This paper introduces a new local variable attribution method for survival models with sound theoretical guarantees like local accuracy. It has been illustrated and validated using synthetically generated data and compared with the SurvLIME method. Moreover, an example of the developed method in a real-life use case is presented. The method's source code, the experiments carried out in this study, and a source code of the so far \emph{not} implemented SurvLIME method are contributed.

SurvSHAP(t) is the first explanation that presents its final results as time-dependent functions. We believe this sets a new direction for research at the intersection of survival analysis and XAI. It is worth pointing out that the approach could be generalized for any model producing functional output. Future works should consider the possibility of aggregating the SurvSHAP(t) function across data distribution to introduce global explanations of machine learning survival models. 

We anticipate that an accessible visualization of SurvSHAP(t) can popularize explainability methods in domains where survival analysis is applied. Our contribution benefits various stakeholders, e.g., physicians and bioinformaticians, in extracting knowledge from data and model analysis. We recommend applying SurvSHAP(t) to explain RSF and deep learning models in scenarios where only the CPH model was previously considered in practice. 

\section*{Acknowledgments}
We would like to thank Mai P. Hoang, MD, from Harvard Medical School, Boston, MA, USA, and Piotr Donizy, MD, from Department of Clinical and Experimental Pathology, Wroclaw Medical University, Wroclaw, Poland, for valuable discussions on the presented method. We also thank Anna Kozak and Katarzyna Woźnica for their valuable comments about the study. This work was financially supported by the NCBiR grant INFOSTRATEG-I/0022/2021-00 and NCN Sonata Bis-9 grant 2019/34/E/ST6/00052.

\bibliographystyle{plainnat}
\bibliography{references} 

\end{document}